\documentclass[lettersize,journal]{IEEEtran}
\usepackage{amsmath,amsfonts}
\usepackage{algorithmic}
\usepackage{algorithm}
\usepackage{array}
\usepackage[caption=false,font=normalsize,labelfont=sf,textfont=sf]{subfig}
\usepackage[colorlinks=true, citecolor=green, linkcolor=red]{hyperref}
\usepackage{textcomp}
\usepackage{stfloats}
\usepackage{url}
\usepackage{verbatim}
\usepackage{graphicx}
\usepackage{cite}
\usepackage{times}
\usepackage{multirow}
\usepackage[misc]{ifsym}
\usepackage[table,xcdraw]{xcolor}   
\usepackage{listings}
\usepackage{arydshln}
\usepackage{booktabs}
\usepackage{pifont}
\usepackage{tcolorbox}
\usepackage{enumitem} 

\newcommand{\eg}{\emph{e.g.}}
\newcommand{\wrt}{\emph{w.r.t.}}

\newcommand{\etc}{\emph{etc. }}

\begin{document}

\title{\textcolor{blue}{RA-SSU}: Towards Fine-Grained Audio-Visual Learning with \textcolor{blue}{R}egion-\textcolor{blue}{A}ware \textcolor{blue}{S}ound \textcolor{blue}{S}ource \textcolor{blue}{U}nderstanding}

\author{Muyi Sun,~\IEEEmembership{Member,~IEEE}, Yixuan Wang, Hong Wang, Chen Su, Man Zhang,~\IEEEmembership{Senior Member,~IEEE}, \\ Xingqun Qi$^{\dagger}$,~\IEEEmembership{Member,~IEEE}, 
Qi Li,~\IEEEmembership{Member,~IEEE}, 
Zhenan Sun,~\IEEEmembership{Senior Member,~IEEE}
\thanks{This work is supported by the BNSF (Grant	24QY0211), the NSFC (Grant 62306309, U23B2054) and the NKRDPC (Grant 2023YFF0904700). (\textbf{Corresponding author: Xingqun Qi.})}
\thanks{Muyi Sun, Yixuan Wang, Hong Wang, Chen Su, and Man Zhang are with the School of Artificial Intelligence, Beijing University of Posts and Telecommunications, Beijing, China. (Email: \{muyi.sun, wyx2021213338, wanghong, suchen1201, zhangman\}@bupt.edu.cn).}
\thanks{Xingqun Qi is with the Academy of Interdisciplinary Studies, The Hong Kong University of Science and Technology, Hong Kong, China. (Email: xingqunqi@gmail.com).}
\thanks{Qi Li and Zhenan Sun are with the Institute of Automation, Chinese Academy of Sciences, and also with the University of Chinese Academy of Sciences, Beijing, China. (Email: qli@nlpr.ia.ac.cn; znsun@nlpr.ia.ac.cn)}
}

\markboth{Acceptance for publication in IEEE Transactions on Multimedia, 8, Feb, 2026.}%
{Shell \MakeLowercase{\textit{et al.}}: A Sample Article Using IEEEtran.cls for IEEE Journals}

\maketitle

\begin{abstract}
Audio-Visual Learning (AVL) is one fundamental task of multi-modality learning and embodied intelligence, displaying the vital role in scene understanding and interaction.
However, previous researchers mostly focus on exploring downstream tasks from a coarse-grained perspective (e.g., audio-visual correspondence, sound source localization, and audio-visual event localization).
Considering providing more specific scene perception details, we newly define a \textbf{fine-grained} Audio-Visual Learning task, termed \textcolor{blue}{\textbf{Region-Aware Sound Source Understanding (RA-SSU)}}, which aims to achieve region-aware, frame-level, and high-quality sound source understanding.
To support this goal, we innovatively construct two corresponding datasets, i.e. fine-grained Music \textcolor{blue}{\textbf{(f-Music)}} and fine-grained Lifescene \textcolor{blue}{\textbf{(f-Lifescene)}}, each containing annotated sound source masks and frame-by-frame textual descriptions.
The f-Music dataset includes 3,976 samples across 22 scene types related to specific application scenarios, focusing on music scenes with complex instrument mixing. 
The f-Lifescene dataset contains 6,156 samples across 61 types representing diverse sounding objects in life scenarios.
Moreover, we propose \textcolor{blue}{SSUFormer}, a \textcolor{blue}{S}ound-\textcolor{blue}{S}ource \textcolor{blue}{U}nderstanding Trans\textcolor{blue}{Former} benchmark that facilitates both the sound source segmentation and sound region description with a multi-modal input and multi-modal output architecture.
Specifically, we design two modules for this framework, \textcolor{blue}{Mask Collaboration Module (MCM}) and \textcolor{blue}{Mixture of Hierarchical-prompted Experts (MoHE)}, to respectively enhance the accuracy and enrich the elaboration of the sound source description.
Extensive experiments are conducted on our two datasets to verify the feasibility of the task, evaluate the availability of the datasets, and demonstrate the superiority of the SSUFormer, which achieves SOTA performance on the Sound Source Understanding benchmark. More details are available on our
project page \url{https://yixuan-wang1.github.io/AVUWEB/}.
\end{abstract}

\section{Introduction}
\label{sec:intro}
In recent years, Audio-Visual Learning (AVL) has become a significant research direction in the field of multimodal learning~\cite{wei2022learning, wei2022perception, zheng2021adversarial}, encompassing various subtasks such as Audio-Visual Correspondence (AVC)~\cite{arandjelovic2017look}, Sound Source Localization (SSL)~\cite{tian2018audio}, and Audio-Visual Event Localization (AVEL)~\cite{feng2023css, xue2021audio}, \etc 
These subtasks aim to uncover the inherent connections within multi-modality data by combining audio and video information, accelerating the exploration of video understanding tasks~\cite{liu2024bavs}.
However, existing AVL tasks are generally limited to holistic or coarse-grained information perception, such as audio-level and video-level category alignment, spatial sound source localization, and temporal event boundary detection.
These coarse-grained approaches have limitations in complex and dynamic real-world scenarios, where finer scene details is essential for perception~\cite{wei2022learning}.
For instance, the AVC focuses on determining the match between audio and video, but it overlooks the spatial localization of the sound-producing objects. 
The SSL task is limited to region-level sound source localization, lacking detailed sound object understanding within scenes. 
The Audio-Visual Event Localization (AVEL) task advances by localizing audio-visual events in time using pre-defined event labels.
However, it remains restricted to coarse-grained event classification and temporal localization, neglecting a deeper understanding of the form and semantic information for sound-producing objects.

\begin{figure*}[h]
  \centering
   \includegraphics[width=0.98\textwidth]{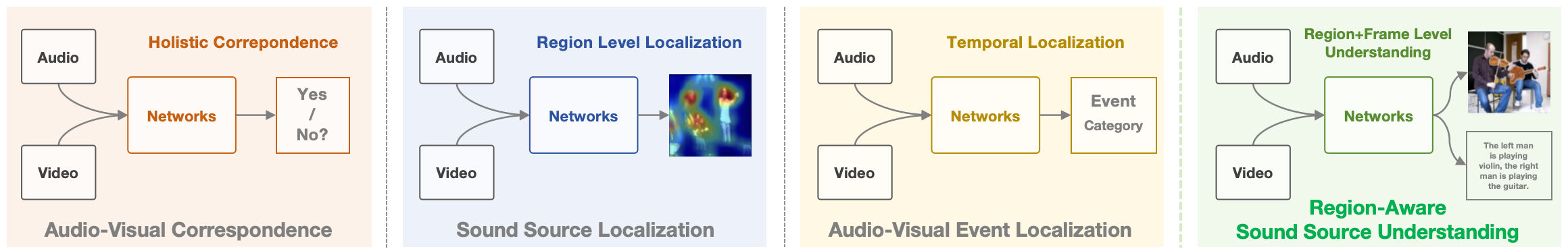}
   \caption{Comparisons of the Region-Aware Sound-Source Understanding task with several previous Audio-Visual Learning tasks. (a).Audio-Visual Correspondence. \textbf{Holistic} correspondence with audio-level and video-level alignment. (b).Sound Source Localization. \textbf{Coarse-grained (spatial)} and region-level sound source localization. (c).Audio-Visual Event Localization. \textbf{Coarse-grained (temporal)} event localization with frame boundaries.  (d). Region-Aware Sound-Source Understanding. \textbf{Fine-grained} audio-visual learning with regional sound source localization and sound region description.}
   \label{Figure1}
\end{figure*}

To address these limitations, in this paper, we propose \textbf{RA-SSU}, a fine-grained audio-visual learning task, which takes the granularities of space and time into account, with \textbf{Region-Aware and frame-level Sound Source Understanding}. 
Unlike previous coarse-grained tasks, RA-SSU is designed for task-specific scenarios, emphasizing not only spatial awareness localization, but also more precise semantic understanding of sound sources, thereby enhancing fine-grained perception capabilities for Sound-Source Understanding in complex situations.
The architecture of the RA-SSU task and its differences compared to previous AVL tasks are illustrated in Fig.~\ref{Figure1}.
Moreover, by focusing on spatial awareness and detailed semantic descriptions of sound-producing objects, the RA-SSU task demonstrates strong potential to enhance the understanding of visual and auditory events. 
For instance, in task like Audio-Visual Retrieval ~\cite{parida2020coordinated, zheng2021adversarial}, RA-SSU could significantly improve the system's ability to identify sounding objects, enabling more precise matching and retrieval of audio-visual content.
In addition, in Audio-Visual Caption~\cite{tian2019audio}, RA-SSU directly enhances the quality of generated descriptions, offering more accurate and detailed semantic information for automatic captioning. 
Therefore, the RA-SSU task provides broader potential and a wider range of application scenarios for downstream audio-visual tasks.

\begin{figure}
\centering
\includegraphics[width=0.98\linewidth]{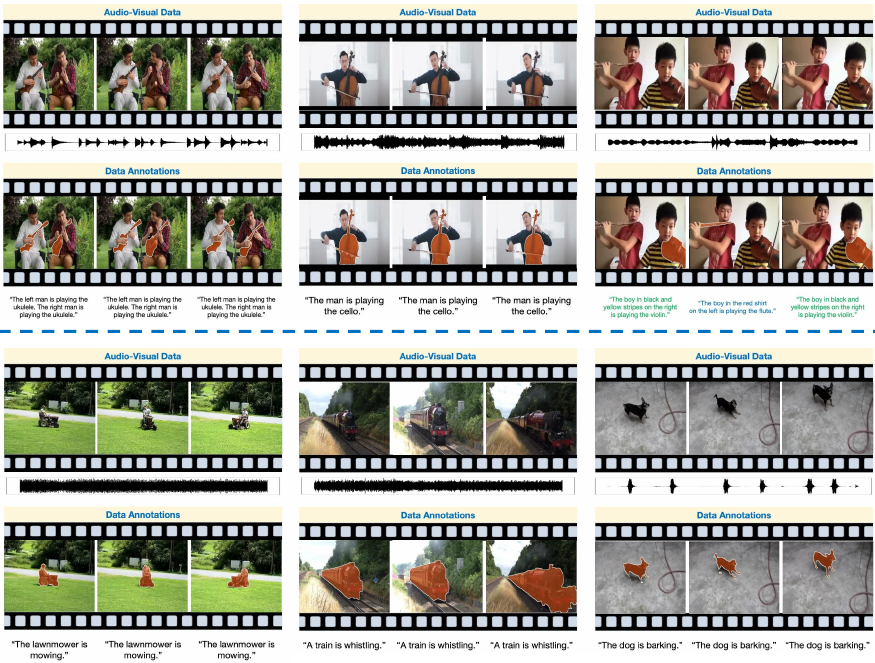} 
\caption{Data samples from the created f-Music and f-Lifescene. 
The top three are from f-Music and down from f-Lifescene.
For each sample, three frames are selected from 10 seconds. 
Each sample contains a 10s video and its frame-level sound source masks and descriptions. Specifically, the frame-by-frame description of a data portion varies with the change of the sounding object, as shown in the upper-right sample. (Zoom in for better details)}
\vspace{-2em}
\label{Figure2}
\end{figure}

To validate the feasibility of the \textbf{RA-SSU} task, we construct two fine-grained audio-visual datasets, \textbf{fine-grained Music (f-Music)} and \textbf{fine-grained Lifescene (f-Lifescene)}. 
The f-Music dataset focuses on music scenes, characterized by complex audio mixtures and background noise. 
In contrast, the f-Lifescene dataset encompasses more intricate everyday scenarios, featuring dynamic interactions among multiple sounding objects.
Compared to music scenes, the everyday scenario dataset presents greater challenges due to its increased scene complexity and semantic richness. 
This complexity necessitates the models effectively manage intricate multi-modality information and achieve nuanced semantic understanding.
These two datasets offer enhanced annotation precision and greater scene diversity, featuring frame-level sound source masks and detailed textual descriptions. 
This rich data support enables the models to train more effectively, improving their performance in complex scenarios compared to existing datasets.
For the size of the datasets, f-Music contains 3976 samples, including 22 types of instrument sound scenes.  F-Lifescene contains 6156 samples, including 61 types of ordinary living environments.
For the data annotation, we utilize the interactive Large Vision Models~\cite{kirillov2023segment} and Large Language Models~\cite{liu2024visual} to obtain the initial rough masks and text description annotations. 
Then, we manually refine the annotation results of the algorithm, resulting in region-aware, frame-level, high-quality masks and descriptions.
The illustrations of the selected data samples are listed in Fig.~\ref{Figure2}.

To achieve the goal of the RA-SSU task, we design a unified framework, named \textbf{SSUFormer}, that accepts both video and audio as inputs, producing segmentation masks for the sounding objects and generating corresponding text descriptions.
To enhance the spatial and semantic consistency of the sound-producing objects, we first propose a Mask Collaboration Module (MCM).
In particular, MCM adaptively leverages the produced masks as spatial guidance to improve the semantic awareness of our framework.
The interactive correlation between two output objectives is modeled by the regional collaborative learning.
In this manner, MCM promotes our framework to yield high-fidelity textual captions with desirable semantic-spatial harmonic properties against the previous single-output counterparts.
Considering the text descriptions should maintain the temporal consistency with the corresponding sequential visual frames, we further propose a Mixture of Hierarchical-prompted Experts (MoHE) module.
Thanks to the advanced LLM techniques, we first utilize the LLaVa~\cite{lin2023video} serving as the linguistics expert to provide long-term temporal representations.
Then, these obtained temporal cues are integrated with frame-wise visual features produced by the transformer-based decoder in a hierarchical router mechanism. 
The temporal-consistent captions are gradually synthesized via the learned routing gate, significantly improving the model's performance, especially when dealing with complex multimodal inputs.
Thus, we build a benchmark that encompasses task definition, data construction, and algorithm development for fine-grained Sound-Source Understanding with a broader range of applications.

In summary, our contributions are summarized as follows:
\noindent
\begin{itemize}[leftmargin=*]
\item We introduce a fine-grained AVL task, named RA-SSU, which goes beyond existing coarse-grained AVL tasks and achieves Region-Aware Sound Source Understanding. 
RA-SSU emphasizes the spatial localization and semantic understanding of sounding objects, enhancing the understanding of complex scenes.
\item To validate the RA-SSU task, we construct two datasets, f-Music and f-Lifescene. 
One focuses on audio mixtures in music scenes and the other covers diverse everyday scenarios, 
which provides higher annotation precision, offering significant data support for fine-grained AVL.
\item We present a multi-modal Transformer backbone, SSUFormer, and design a Mask Collaboration Module with Regional Consistency, propose a Mixture of Hierarchical-prompted Experts Module, which cooperate with each other and jointly improve the SSU performance. 
\item Extensive experiments are conducted to demonstrate the superiority of the SSUFormer, which achieves SOTA on the RA-SSU benchmark and also shows unique advantages compared to multi-modal large models.
\end{itemize}

\section{Related Work}
\label{sec:related_work}

\subsection{Audio-Visual Learning Tasks and Datasets}
In recent years, Audio-Visual Learning~\cite{xu2024each, shi2024cross, chen2022audio} has aimed to achieve human-like perception by exploring the synergetic integration of auditory and visual information. 
Arandjelovic et al.~\cite{arandjelovic2018objects} propose a framework for cross-modal retrieval and sound source localization, leveraging unlabeled videos to perform self-supervised learning for the audio-visual correspondence (AVC). 
Similarly, Owens et al.~\cite{owens2018audio} emphasize the fusion of multisensory representations, training networks in a self-supervised manner to predict the temporal alignment between audio and video frames.
Senocak et al.~\cite{senocak2018learning} design a dual-stream network, combining audio and visual information using unsupervised learning and attention mechanisms to achieve sound source localization. 
Meanwhile, Zhao et al.~\cite{zhao2018sound} learn to locate sound-producing regions within images and separate sounds into components corresponding to individual pixels by processing unlabeled videos. 
Wang et al.~\cite{wang2024ref} propose a referring audio-visual segmentation task to refer and segment objects in audio-visual scenes.
Zhou et al.~\cite{zhou2022audio} introduce the Audio-Visual Segmentation (AVS) task to generate pixel-level maps of sound-producing objects in image frames.
Zhou et al.~\cite{zhou2025dense} propose the Dense Audiovisual Event Localization (DAVEL) task, which aims to identify all audiovisual events in uncropped videos and provide precise frame-level temporal boundaries. Besides, to address the challenge of distinguishing between audio, visual, and audiovisual events in the Audiovisual Video Parsing (AVVP) task, Zhou et al.~\cite{zhao2025multimodal} propose a Multimodal Category-Aware Semantic Enhancement (MCSE) network, introducing event category semantics to enhance feature representations.
These works reflect a progression in the field of audio-visual perception, evolving from the region-level localization of sound-producing objects in SSL to the pixel-level segmentation in AVS. 
Huan et al.\cite{xuan2022proposal} pioneer proposal-based MIL with an audio-conditioned GRM, eliminating manual boxes and boosting IoU over heat-map baselines. 
Meanwhile, the extended work~\cite{xuan2024robust} couples proposal detection with contrastive learning, using Active Contrast-Set Mining and GRM re-weighting to learn noise-robust object-level sound sources.
This trend demonstrates a pursuit of a fine-grained perception of both audio and video. 

However, much of the existing research focuses on specific tasks and concentrates on holistic or coarse-grained perception, overlooking interactions with other elements in the scene and the semantic understanding of the environment.
This results in a lack of comprehensive scene understanding, as seen in~\cite{tian2018audio}, which attempts to precisely localize audio-visual events along the timeline. 
CMANet\cite{xuan2020cross} aligns out-of-sync audio-visual events via stacked spatial–temporal– modal attention. DCMANet\cite{xuan2021discriminative} refines this with adaptive DCMAN attention and
an eigenvalue regularizer, attaining AVEL task state-of-the-art under full and weak supervision.
AVEL identifies visible and audible events in videos through predefined event labels, yet it remains primarily focused on temporal event classification, neglecting a detailed understanding of the form and semantics of the sound-producing objects. 
This limitation becomes particularly evident in complex scenarios.
To address this, we propose a fine-grained task-specific AVL task, RA-SSU, which not only emphasizes regional awareness but also captures the spatial location and semantic information of sound objects. 

For the audio-visual datasets, there are also multiple datasets which have been appropriately labeled for different tasks, such as audio-visual speech recognition~\cite{patterson2002moving, afouras2018deep, QI2026103613}, action recognition~\cite{chung2018voxceleb2, duan2025article}, emotion recognition~\cite{zadeh2017tensor, qi2024emotiongesture, qi2024weakly}, speech separation~\cite{harte2015tcd}, event localization~\cite{chatterjee2021visual}, question answering~\cite{lei2018tvqa}.
Please refer to Tab.~\ref{tab:comparison} for specific tasks and annotation contents of the datasets.
Similar to the above task comparisons, we have newly constructed our own task data while referring to these datasets. 
We focus on fine-grained perception tasks and annotate frame-level and region-aware masks and descriptions.

\begin{table}[t]
\centering
\caption{The statistics of representative audio-visual datasets.}
\label{tab:comparison}
\resizebox{0.46\textwidth}{!}{
\begin{tabular}{lcc}
\toprule
  \textbf{Datasets} &\textbf{Annotations} &\textbf{Tasks}\\ \midrule \midrule
CUAVE~\cite{patterson2002moving} &
  \begin{tabular}[c]{@{}c@{}}Vocabulary\\ Lip Edge\end{tabular} &
  \begin{tabular}[c]{@{}c@{}}Audio-Visual \\ Speech Recognition\end{tabular} \\ \midrule
LRS2-BBC~\cite{afouras2018deep} &
  \begin{tabular}[c]{@{}c@{}}Speech Text\\ Lip Movement\end{tabular} &
  \begin{tabular}[c]{@{}c@{}}Audio-Visual \\ Speech Recognition\end{tabular} \\ \midrule
VGG-Sound~\cite{chen2020vggsound} &
  Scene Category &
  \begin{tabular}[c]{@{}c@{}}Audio-Visual \\ Correspondence\end{tabular} \\ \midrule
VoxCeleb1/2{~\cite{chung2018voxceleb2}} &
  \begin{tabular}[c]{@{}c@{}}Facial Trajectory\\ Speaker ID\end{tabular} &
  \begin{tabular}[c]{@{}c@{}}Audio-Visual \\ Speaker Recognition\end{tabular} \\ \midrule
Hollywood~\cite{laptev2008learning} &
  \begin{tabular}[c]{@{}c@{}}Action Category\\ Video-level Caption\end{tabular} &
  \begin{tabular}[c]{@{}c@{}}Audio-Visual \\ Action Recognition\end{tabular} \\ \midrule
Kinetics{~\cite{kay2017kinetics}} &
  \begin{tabular}[c]{@{}c@{}}Action Category\\ Action Localization\end{tabular} &
  \begin{tabular}[c]{@{}c@{}}Audio-Visual \\ Action Recognition\end{tabular} \\ \midrule
CMU-MOSI{~\cite{zadeh2017tensor}} &
  Emotion Category &
  \begin{tabular}[c]{@{}c@{}}Audio-Visual \\ Emotion Recognition\end{tabular} \\ \midrule
MaSaC{~\cite{bedi2021multi}} &
  Emotion Category &
  \begin{tabular}[c]{@{}c@{}}Audio-Visual \\ Emotion Recognition\end{tabular} \\ \midrule
TCD-TIMIT{~\cite{harte2015tcd}} &
  \begin{tabular}[c]{@{}c@{}}Speech Text\\ Speaker ID\end{tabular} &
  \begin{tabular}[c]{@{}c@{}}Audio-Visual \\ Speech Enhancement\\ Speech Separation\end{tabular} \\ \midrule
Ref-AVS~\cite{wang2024ref} &
  \begin{tabular}[c]{@{}c@{}}Types Masks\\ Expression\end{tabular} &
  \begin{tabular}[c]{@{}c@{}}Audio-Visual  \\ Referring Segmentation\end{tabular} \\ \midrule
MUSIC~\cite{zhao2018sound} &
  Masks &
  \begin{tabular}[c]{@{}c@{}}Object Sound Separation \\ Sound Source Localization\end{tabular} \\ \midrule
ASIW~\cite{chatterjee2021visual} &
  \begin{tabular}[c]{@{}c@{}}Event Categories\\ Captions\end{tabular} &
  \begin{tabular}[c]{@{}c@{}}Object Sound Separation \\ Event Localization\end{tabular} \\ \midrule
MusicAVQA~\cite{lei2018tvqa} &
  QA Pairs &
  \begin{tabular}[c]{@{}c@{}}Audio-Visual  \\ Question Answering\end{tabular} \\ \midrule
  \rowcolor[HTML]{ECF4FF} 
\textbf{Ours} &
  \begin{tabular}[c]{@{}c@{}}\textbf{Text Descriptions}\\ \textbf{Masks}\end{tabular} &
  \begin{tabular}[c]{@{}c@{}}\textbf{Region-Aware Sound} \\ \textbf{Source Understanding}\end{tabular} \\ \bottomrule
\end{tabular}}
\end{table}

\subsection{Video Understanding}

Video understanding has made significant progress in recent years, with increasing attention focused on various tasks such as video classification and action recognition~\cite{soomro2012ucf101,kuehne2011hmdb,li2025vividlistener, zhao2025freedance, zhang2025danceeditor},  temporal action localization~\cite{cheng2022tallformer,zhang2022actionformer,wang2022rcl,hu2025remerec}, and video question-answering~\cite{zellers2019recognition,xu2017video}. 
These diverse tasks have showcased improved performance in video understanding. Video caption tasks encompass both standard video captioning and dense video captioning. Standard video caption is a key task in video understanding, where coherent textual descriptions of video content are generated. Common datasets used for this task include MSRVTT~\cite{xu2016msr}, YOUCOOK2~\cite{zhou2018towards}, and MSVD~\cite{chen2011collecting}, with evaluation metrics such as BLEU~\cite{papineni2002bleu}, METEOR~\cite{banerjee2005meteor}, ROUGE~\cite{lin2004rouge}, and CIDEr~\cite{vedantam2015cider}. 
With advancements in technology, researchers have further explored the task of dense video captioning, which aims to generate detailed textual descriptions that cover multiple events within a video, requiring a deeper understanding of spatio-temporal information.
Moreover, the rapid growth in video understanding is closely linked to the development of large language models (LLMs), which have also been widely applied in video understanding. By integrating visual encoders with LLMs~\cite{zhang2023video}, models are able to handle video-to-text decoding tasks, enabling multi-modality video analysis. 
In this paper, we focus on a fine-grained audio-visual learning task to empower video understanding with a more detailed perception.

\section{Preliminaries}

\subsection{Task Definition}
\begin{figure}[t]
  \centering
   \includegraphics[width=0.49\textwidth]{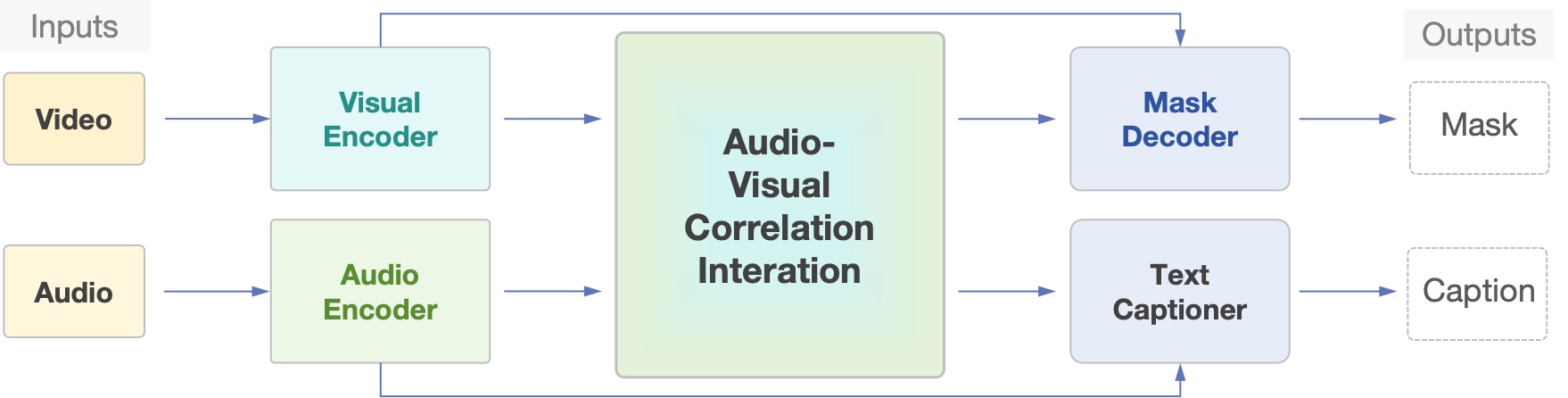}
   \caption{The framework of the RA-SSU task. Multi-modal inputs are first processed by modality-specific encoders. 
   Then the main networks are designed to align, interact, and integrate audio-visual features. Task Decoders realize the mask and description prediction combined with the initial representations.}
   \label{Figure3}
\end{figure}

In this paper, we propose the Region-Aware Sound-Source Understanding task. 
Our goal is to achieve fine-grained, region-aware, and frame-level video understanding.
Given a 10-second video, we first extract the audio signal and resample the video frames to standard 30fps.
Then, we employ the powerful pre-trained visual and audio encoders to obtain aligned embeddings.
With the region-aware annotation masks and the text descriptions, the main goal is to design a mapping algorithm with audio/video as input, exporting sound object masks and descriptions to approximate the target annotated ones. 
Thus, the task can be defined with the formula:
\begin{align}
(M^p, T^p) = Network(Enc_V(V),Enc_A(A)). 
\tag{1}
\end{align}
Where the $V$ and $A$ denote the extracted audio signals and resampled visual frames, respectively. $Enc_V(\cdot)$ and $Enc_A(\cdot)$ indicate the corresponding visual and audio encoders.
The $Network$ acts as a multi-modal input and multi-modal output Transformer architecture with specific decoders. $M$ means semantic mask and $T$ represents text descriptions, $p$ means network prediction.
It effectively integrates audio and video information to generate region masks and text embeddings. These outputs are crucial for enhancing sound source understanding, allowing for a more nuanced interpretation of audio-visual content.
The total task pipeline is illustrated in Fig.~\ref{Figure3}.

\subsection{Dataset Construction}
\subsubsection{Dataset Collection and Protocols}
Inspired by the previous research related to audio-visual datasets\cite{chen2020vggsound,li2022learning}, we attempt to build databases to verify the sound-source understanding task from two perspectives: specific small scenes and complex life scenarios. 
Both datasets feature dynamic scenarios that include either a single sound source or a mixture of complex sound sources. This diversity provides a comprehensive foundation for advancing audio-visual learning.

For the specific small scenes, we focus on the music scenes, since almost all the sounding objects come from various types of musical instrument performances. 
Thus, inherited from the Music-AVQA dataset\cite{li2022learning}, we build a fine-grained Music dataset \textbf{f-Music}.
Unlike Music-AVQA, which is primarily designed for open-ended video question-answering tasks, f-Music focuses specifically on video data featuring spatially sensitive and temporally dynamic musical instruments. 
This distinction allows f-Music to provide a more nuanced analysis of audio-visual interactions in musical contexts.

To address complex life scenarios, we have developed f-Lifescene, utilizing the audio-visual dataset VGG-Sound~\cite{chen2020vggsound}, which comprises general audio-visual data extracted from YouTube videos. 
This comprehensive categorization allows f-Lifescene to encompass a wide range of real-life scenarios relevant to our RA-SSU tasks.
In short, we have built two datasets, one from a small perspective and one from a complex and comprehensive perspective, to validate our proposed task.

\subsubsection{Dataset Pre-processing}
In the data collection procedure, we use the keywords of thr subcategories to retrieve videos from the initial data. 
For the single-sound source data, we sample 50 to 100 videos for each keyword, clipping them to 10 seconds at 30 frames per second (fps). The image resolution of these videos is uniformly constrained to 512$\times$512.
For the multi-source data, we combine the multi-keywords to retrieve related videos. 
We manually assess the retrieval accuracy of the dataset, as well as the matching accuracy between audio and video.
To enhance the diversity of the multi-source data, we employ 
audio overlay and video stitching operations. 


\begin{figure}[h]
  \centering
   \includegraphics[width=0.99\linewidth]{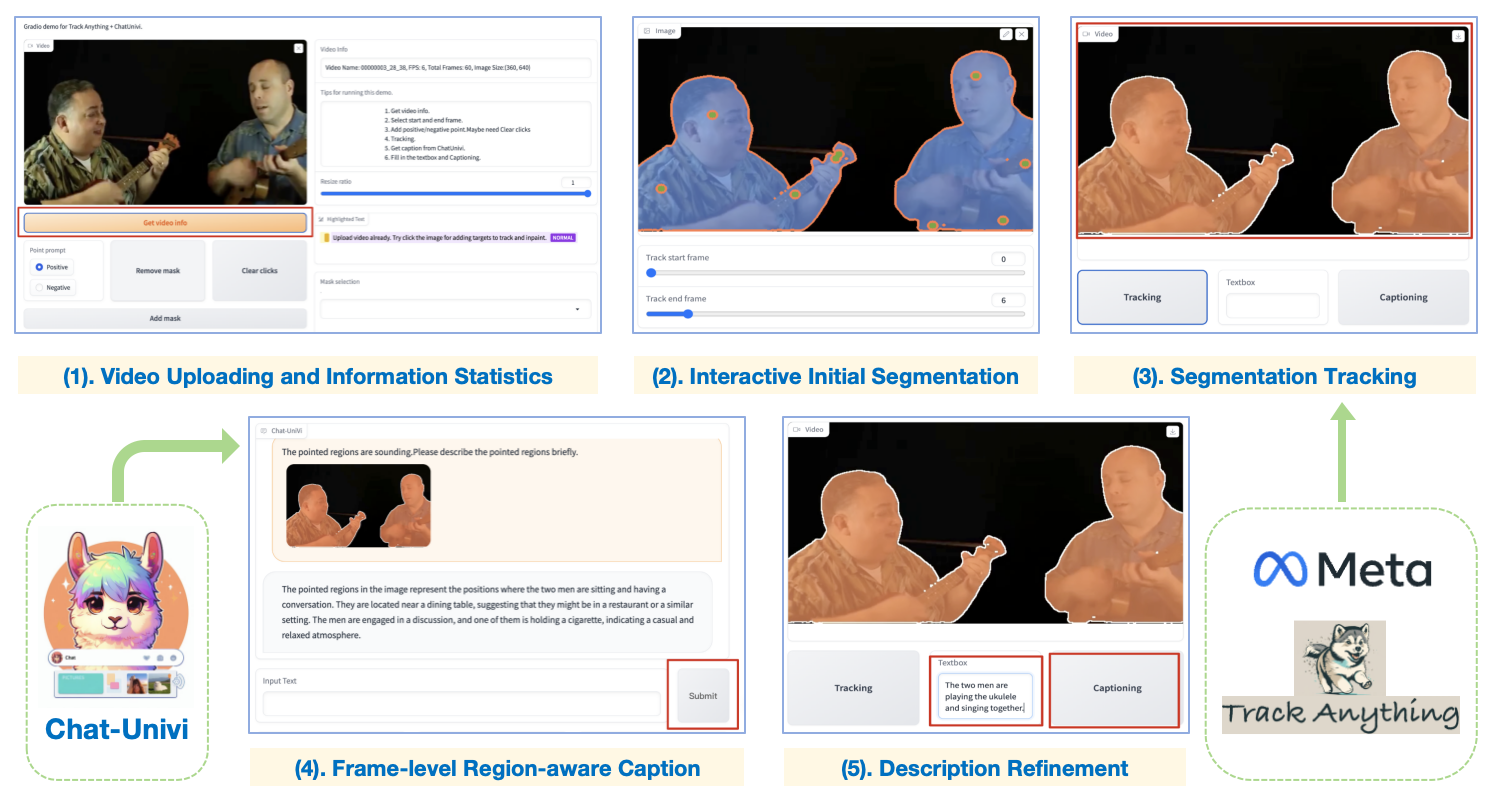}
   \caption{Data annotation process and labeling system for the proposed sound-source understanding task. In this system, the video data is first uploaded into the system. Then the SAM model is used to obtain the initial masks. Based on the initial masks, the TAM is used to gain the frame-level video masks. Finally, the masked region of the frame-level images is fed into the Chat-Univi to get the region-aware descriptions. (Zoom in for better details.)}
   \label{Figure5}
\end{figure}

\subsubsection{Dataset Annotation}
For Sound-Source Understanding, the corresponding datasets need to obtain the region-aware masks and descriptions.
Manually processing each frame of nearly 10,000 video samples would involve a significant workload. 
However, advancements in Large Vision Models~\cite{zhao2023fast} and Visual Language Models~\cite{dong2023maskclip, lin2024vila} provided open-source model interfaces that pave the path for data annotation. Notable examples include the Segment Anything Model (SAM)~\cite{kirillov2023segment}, Track Anything Model (TAM)~\cite{yang2023track}, and LLaVA~\cite{liu2024visual}. Inspired by these methods, we design a novel labeling strategy and construct an efficient labeling system based on human-algorithm interaction\footnote{For more details about data collection, pre-processing, and statistics analysis, please refer to the supplementary material.}. The data annotation process and system are shown in Fig.~\ref{Figure5}.



\section{Methodology}

\begin{figure*}[h]
  \centering
   \includegraphics[width=1\textwidth]{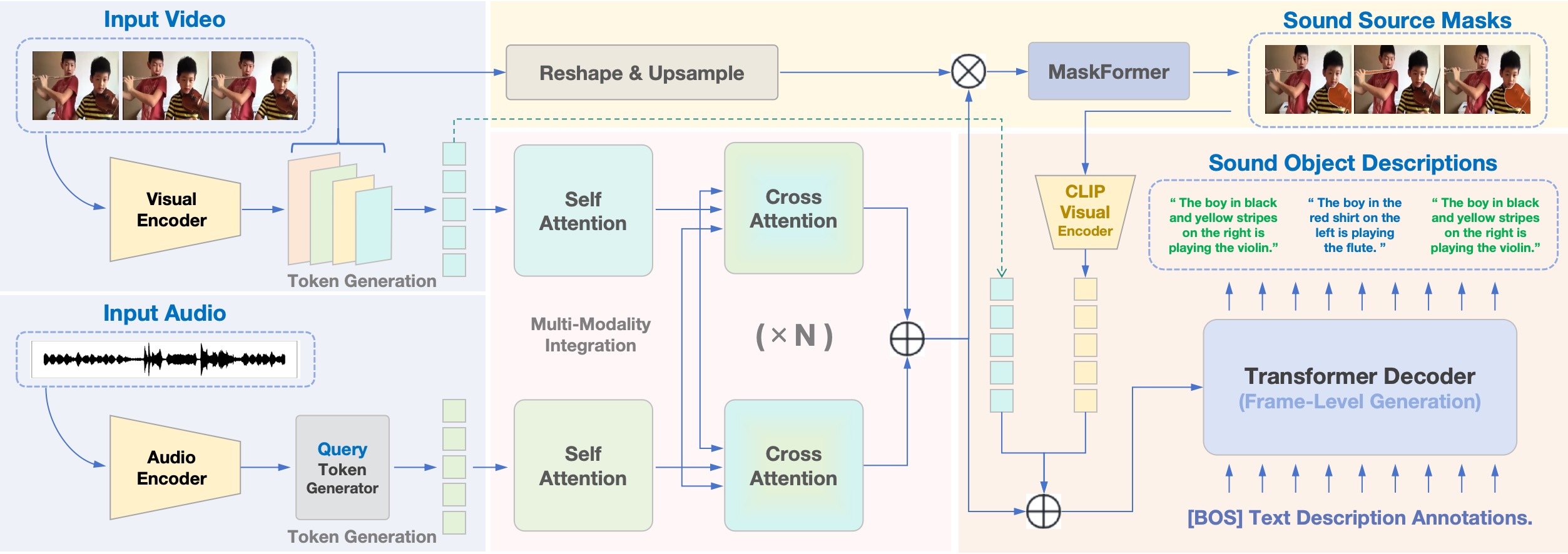}
   \caption{SSUFormer: fine-grained Sound-Source Understanding Benchmark. On the left of this architecture, the audio and video are fed into the encoders and mapped to the token representations. 
   Then, the multi-modality features are fused with the attention mechanism in Fig.6. Next, the previous features are integrated into task decoders for mask and description generation. (Zoom in for better details.)}
   \label{Figure6}
   \vspace{-2mm}
\end{figure*}

\subsection{Overview Framework}\label{subsec2}
As described in the section on Task Definition and Fig.~\ref{Figure3}, the total framework is a multi-modality input and multi-modality output architecture, which builds a bridge for multi-modality representation.
For input, standardized or normalized video and audio are fed into their respective encoders.
These two encoders are generally single-modality pre-trained models.  
We employ the VGGish~\cite{hershey2017cnn} as the audio encoder, a specific neural network for the extraction and classification of audio features.
Similarly, we employ PVT-v2~\cite{wang2022pvt} as the video encoder to extract the visual representations.
In the Audio-Visual Correlation Interaction phase, we design cross-attention modules for multimodal integration. 
Based on the integrated multimodal features, we propose task decoders for mask and description generation, incorporating a collaborative strategy to facilitate mutual enhancement between the two tasks. 
This collaborative strategy is implemented within the Mask Collaboration Module.
The main framework is built on the Transformer architecture, with the mask generation decoder following the MaskFormer~\cite{cheng2022masked} and the description generation decoder adhering to an auto-regressive Transformer.

\subsection{SSUFormer: Sound-Source Understanding Benchmark}
As shown in Fig.~\ref{Figure6}, the total Sound-Source Understanding Transformer (SSUFormer) comprises four parts: Data Input and Feature Embedding Part, Multi-Modality Integration Part, Sound Source Segmentation Part, and Sound Region Description Part. 
The attention mechanism is employed in the Multi-Modality Integration Part.
The Mask Collaboration Module is fused into the Sound Region Description Part.
The following provides a detailed explanation of each part.

\subsubsection{Data Input and Feature Embedding Part} 
In this part, the audio and video are fed into the specific encoders.
We employ VGGish~\cite{hershey2017cnn} as the audio encoder and PVT-v2~\cite{wang2022pvt} as the video encoder. VGGish is a simple yet effective convolutional neural network specifically designed for large-scale audio event classification, with pre-training conducted on the AudioSet~\cite{gemmeke2017audio}.
PVT-v2 serves as an improved baseline for the Pyramid Vision Transformer, offering a lightweight and straightforward design for integration. It is pre-trained on the ImageNet-1K dataset~\cite{russakovsky2015imagenet}.
We chose these two methods to ensure low computational consumption while maintaining strong feature representation capabilities. The focus of this work is to implement an audio and video interaction and fusion model, where a basic and effective representation of input data is sufficient.

\subsubsection{Multi-Modality Integration Part} 
In this part, the tokenized audio and video features are integrated for subsequent predictions.
We employ two traditional strategies from the Transformer architecture for feature learning: self-attention and cross-attention mechanisms.
Within these attention modules, three learnable projection matrices are shared across both spatial and temporal perceptions. 
Then, we leverage the middle audio features as the query to match the key features and value features, which are acted by the middle video embeddings.
In this fashion, the semantic information from the visual frames and the temporal representation are adaptively aligned to produce the infused joint features. The learning procedure is expressed as:
\begin{align}
    Q =W_{Q} \cdot f_{a}, K=W_{K} \cdot F_{v}, V=W_{V} \cdot F_{v},
    \tag{2}
\label{10}
\end{align}
where the $W_{Q}$, $W_{K}$, $W_{V}$ denote learnable projection matrices, $f_{a}$ is the middle audio features, $F_{v}$ is the visual features. Once we acquire the learned query, key, and value embedding, we utilize a final softmax operation to produce the output of Multi-Modality Integration (MMI), formulated as:
\begin{align}
    F_{att} & =MMI(Q, K, V),
    \tag{3}
\end{align}
where $F_{att}$ is the infused features guided by visual-audio modality.
With a similar structure, the self-attention module processes single-modality information, while the cross-attention module handles audio-visual information. This design enhances and fuses cross-modality features. 
We conduct attention structure N(=3) times in the total framework.

\subsubsection{Sound Source Segmentation Part} 
In this part, a MaskFormer~\cite{cheng2022masked} based architecture is employed for mask prediction. 
MaskFormer is a segmentation method with powerful semantic parsing ability that can also be applied to audio-visual segmentation, using audio features as queries. 
The final sound source masks are generated by a multi-layer perceptron (MLP) for pixel-level prediction. Consequently, object-level regions are produced, which can be interpreted as fine-grained, frame-level spatial localization.

\subsubsection{Sound Region Description Part} 
Inspired by the GPT and BERT~\cite{kenton2019bert}, we design a straightforward Transformer Decoder for text description generation. 
Unlike large language models trained with large-scale datasets, we train this small auto-regressive language model on our task-specific audio-visual annotations, with the fused audio-visual features as the conditions.  
Thus, the sound region description decoder is constructed.

\textbf{Mask Collaboration Module.} To enhance region-aware descriptions, we design a Mask Collaboration Module for improved region-aligned text generation. Our goal is to obtain accurate sound source descriptions, and providing more spatial perception information can theoretically yield better results. Therefore, we treat the masked image areas from the sound source segmentation as input for the text decoder. 
Meanwhile, to directly align the text representation with the regional image, we employ the CLIP~\cite{chen2023vlp} visual encoder for token generation, as illustrated in Fig.~\ref{Figure8} (b). 
This module enables caption results to transition from image-level to region-level granularity, aligning more closely with our initial task objectives.
Moreover, by increasing these interactions, the text description task also enhances segmentation performance, forming a task-collaboration mechanism supported by CLIP Loss (Regional Consistency Constraint) in the subsequent steps.


\begin{figure}
  \centering
   \includegraphics[width=0.98\linewidth]{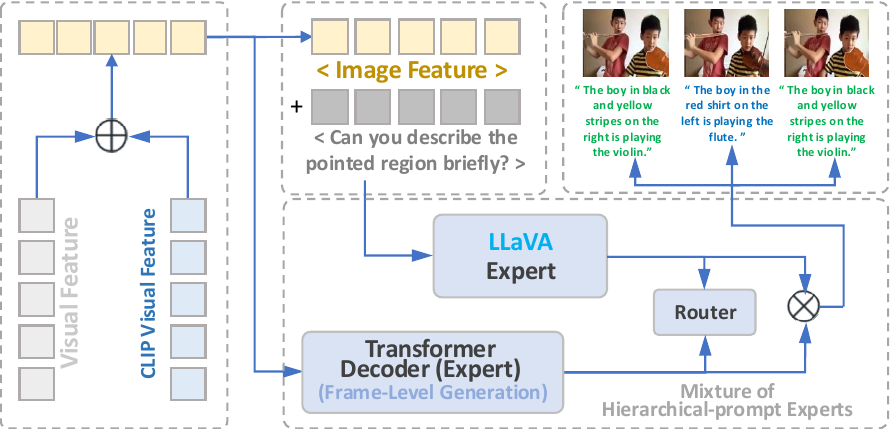}
   \caption{Illustration of proposed Mixture-of-Hierarchical-prompted Experts (MoHE). The MoHE adaptively integrates the text embeddings and frame-wise visual features via a router to produce temporal consistency captions in the long sequence. (Zoom in for better details.)}
   \label{MoHE}
   \vspace{-1em}
\end{figure}

\begin{figure*}[h]
  \centering
   \includegraphics[width=0.95\textwidth]{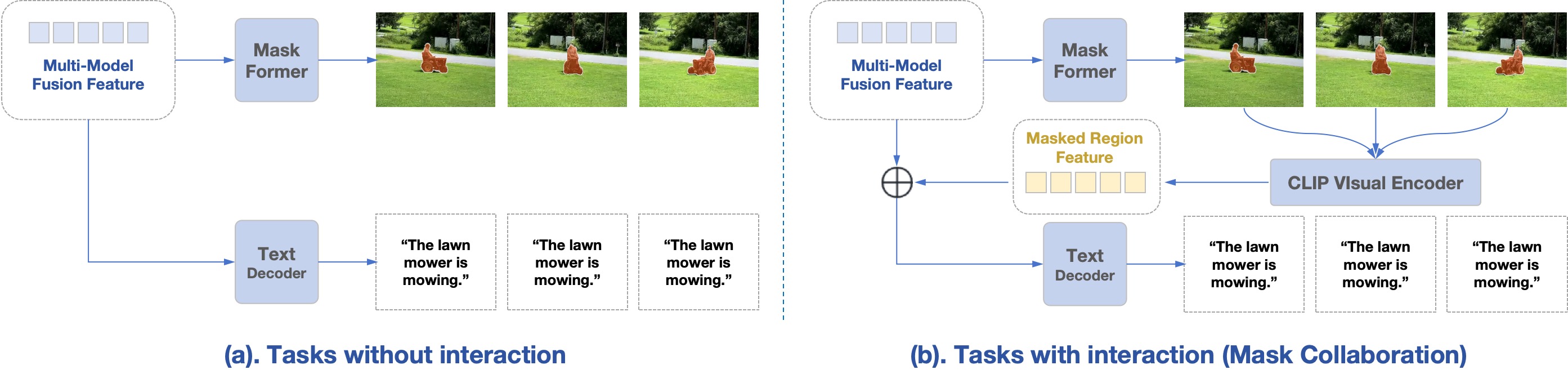}
   \caption{Mask Collaboration Module for task interaction. (a). The plain multi-task output without interaction.
   (b). Mask collaboration module introduces the interaction between these two tasks.
   As a fine-grained RA-SSU task, region-aware visual information will give more details for the text description.
   Thus, combining the multi-modality representations with more regional visual features will give more accurate fine-grained captions.}
   \label{Figure8}
   \vspace{-4mm}
\end{figure*}

\textbf{Mixture of Hierarchical-prompted Experts.} 
Beyond the benchmark architecture in Fig.~\ref{Figure6}, we further propose an improved Mixture of Experts (MoE) module, named Mixture of Hierarchical-prompted Experts (MoHE), which is connected at the end of the benchmark.
The MoHE integrates the LLaVa\cite{lin2023video} Expert with the caption decoder to achieve elaborative description generation as shown in Fig.~\ref{MoHE}. 
In particular, we leverage a router to adaptively balance the dependence weights between the \textbf{visual-text combined} prompt embeddings for visual-language model LLaVA and the total Transformer decoder features, which could be treated as hierarchical prompt prediction (large model + specific decoder). 
In such a manner, the produced captions display coherent alignment with visual frames while preserving temporal consistency in long-sequence modeling.


\subsection{Objective Functions}
As a baseline model for the RA-SSU benchmark, we employ several basic and effective constraints tailored to the specific tasks. Additionally, we introduce a new supervision called the Regional Consistency Constraint (RCC), which enhances the existing network architecture.

Specifically, we use Dice loss $\mathcal{L}_{seg}$ as the constraint for the mask prediction, Focal loss $\mathcal{L}_{tex}$ as the auto-regressive learning function for the text description generation.
To dynamically combine these two losses into a total task loss, we implement an adaptive learning strategy with a hyper-parameter $a$, expressed as:
\begin{align}
\mathcal{L}_{ada}= a\mathcal{L}_{seg} +(1-a)\mathcal{L}_{tex},
\tag{4}
\end{align}
where we set $a=0.25$ in practice.

\textbf{Regional Consistency Constraint.}
Furthermore, to enhance the ability of region-aware perception, we propose a regional consistency loss, which aligns the visual features and textual description features of the sound object. 
Specifically, for the two output parts of SSUFormer, our goal is to obtain the frame-level and fine-grained spatial perception and continuous description of the sounding object.
Therefore, our regional consistency constraint is built upon the insight that providing supervision to boost the coherent consistency between the spatial-wise visual representation and temporal-wise text description embeddings.
Here, we utilize the visual encoder from MaskCLIP~\cite{dong2023maskclip} to encode the output masked regions, while the text encoder processes the textual description features. Consistency constraints are then applied to both encoded outputs, displayed as: 
\begin{align}
\mathcal{L}_{clip} = ||MaskCLIP_v(M^p) - MaskCLIP_t(T^p)||_2,
\tag{5}
\end{align}
where $MaskCLIP_v$ refers to the regional visual encoder and $MaskCLIP_t$ is the text encoder.
The overall objective $\mathcal{L}_{ssu}$ of our model is represented as:
\begin{align}
\mathcal{L}_{ssu}= \mathcal{L}_{ada} + \mathcal{L}_{clip}.
\tag{6}
\end{align}

\begin{table}[t]
\centering
\caption{Quantitative comparisons with single-task models on f-Music. SSS: Sound Source Segmentation, AVC: Audio-Visual Caption, SSU-D: Sound-Source Understanding with Description Only, SSU-S: Sound-Source Understanding with Segmentation Only.}
\label{tab:tab2}
\setlength{\tabcolsep}{3 mm}
\resizebox{0.48\textwidth}{!}{
\begin{tabular}{lccccc}
\toprule
\multicolumn{6}{c}{\textbf{f-Music}} \\ \midrule \midrule
Methods & mIoU & F-score & BLEU & ROUGE-L & METEOR \\ \midrule
SSS~\cite{gao2024avsegformer}\textcolor[HTML]{C0C0C0}{$_{AAAI'24}$} & 0.5919 & 0.7597 & - & - & - \\
AVSBench~\cite{zhou2025audio}\textcolor[HTML]{C0C0C0}{$_{IJCV'25}$} & 0.5663 & 0.7191 & - & - & - \\
AVIS~\cite{guo2025audio}\textcolor[HTML]{C0C0C0}{$_{CVPR'25}$} & 0.5968 & 0.7227 & - & - & - \\
SSU-S & 0.5987 & 0.7552 & - & - & - \\
AVC\cite{tian2019audio}\textcolor[HTML]{C0C0C0}{$_{CVPR'19}$} & - & - & 0.2546 & 0.3650 & 0.2668 \\
SSU-D & - & - & 0.3558 & 0.4497 & 0.3380 \\ 
PG-Video-LLaVA~\cite{munasinghe2023pg} & 0.5554 & 0.7393 & 0.2701 & 0.3134 & 0.2147 \\
\midrule
\rowcolor[HTML]{ECF4FF} 
\textbf{SSU (ours)} & \textbf{0.6280} & \textbf{0.7792} & \textbf{0.4049} & \textbf{0.4882} & \textbf{0.4002} \\ \bottomrule
\end{tabular}%
}
\end{table}

\begin{table}[t]
\centering
\caption{Quantitative comparisons with single-task models on f-Lifescene. (Abbreviation similar to TABLE. III).}
\label{tab:tab3}
\setlength{\tabcolsep}{3 mm}
\resizebox{0.48\textwidth}{!}{
\begin{tabular}{lccccc}
\toprule
\multicolumn{6}{c}{\textbf{f-Lifescene}} \\ \midrule \midrule
Methods & mIoU & F-score & BLEU & ROUGE-L & METEOR \\ \midrule
SSS~\cite{gao2024avsegformer}\textcolor[HTML]{C0C0C0}{$_{AAAI'24}$} & 0.6340 & 0.7064  & - & - & - \\
AVIS~\cite{guo2025audio}\textcolor[HTML]{C0C0C0}{$_{CVPR'25}$} & 0.5968 & 0.5697 & - & - & - \\
AVSBench~\cite{zhou2025audio}\textcolor[HTML]{C0C0C0}{$_{IJCV'25}$} & 0.5697 & 0.6029  & - & - & - \\
SSU-S & 0.6320 & 0.7256 & - & - & - \\
AVC\cite{tian2019audio}\textcolor[HTML]{C0C0C0}{$_{CVPR'19}$} & - & - & 0.3590 & 0.3423 & 0.3618 \\
SSU-D & - & - & 0.5169 & 0.5025 & 0.4815 \\
PG-Video-LLaVA~\cite{munasinghe2023pg} & 0.5134 & 0.5676 & 0.1900 & 0.2164 & 0.2374 \\
\midrule
\rowcolor[HTML]{ECF4FF} 
\textbf{SSU (ours)} & \textbf{0.6530} & \textbf{0.7368} & \textbf{0.5290} & \textbf{0.5388} & \textbf{0.5050} \\ \bottomrule
\end{tabular}%
}
\vspace{-1em}
\end{table}

\section{Experiments}
In this section, we systematically introduce the experimental verification of SSUFormer on f-Music and f-Lifescene. We cover the training details, evaluation metrics, module ablation analysis, and both quantitative and qualitative comparisons with other audio-visual methods and multi-modal large models addressing similar tasks.
Furthermore, we discuss the emergence of the proposed RA-SSU task, highlighting its unique advantages. 
Additionally, we verify that the produced fine-grained perception enhances existing tasks such as audio-visual retrieval and video captioning.

\textbf{Implementation Details.} 
The SSUFormer model is implemented by PyTorch with 4 NVIDIA V100 GPUs under a batch size of 4. We use Adam optimizer as the gradient descent method and set the learning rate $\alpha=0.00002$. The model is trained with 60 epochs for both datasets. 
Based on the size of the datasets, we divide 75\%, 10\%, and 15\% of the f-Music dataset and 70\%, 15\%, and 15\% of the f-Lifescene dataset into training, validation, and testing sets. 
All the categories in the testing set are balanced to better validate the generalization of the model.
For the adaptive loss, we use a random initialization strategy to set the hyperparameter.

\textbf{Evaluation Metrics.} We use mIoU and F-score for segmentation evaluation~\cite{gao2024avsegformer}, and BLEU~\cite{papineni2002bleu}, ROUGE-L~\cite{lin2004rouge}, and METEOR~\cite{banerjee2005meteor} to assess the performance of text generation. 
mIoU measures the overlap between the predicted results and the ground truth, which can clearly reflect the accuracy of the model in segmentation tasks. 
F-score takes into account both precision and recall, making it suitable for imbalanced datasets, since there are still differences in the number of categories. 
BLEU is used to measure the similarity between the generated text and the reference text based on n-grams, and it can automatically assess the fluency and accuracy of the generated text. 
ROUGE-L captures the sequential information between sentences, making it suitable for evaluating the quality of long text generation. 
Compared to BLEU, METEOR has stronger semantic matching, making it more appropriate to assess semantic retention in translation tasks.

\subsection{Quantitative Analysis}
\subsubsection{Comparisons with Single-task Models}
In this paper, we first compare the SSUFormer with previous single-task audio-visual models.
To validate our method, we choose suitable open-source models to implement sound source segmentation~\cite{gao2024avsegformer, guo2025audio, 
zhou2025audio} and audio-visual caption~\cite{tian2019audio}. 
Meanwhile, we utilize our benchmark to achieve single-task and final dual-task training. The performance of these two datasets is shown in Tab.~\ref{tab:tab2} and Tab.~\ref{tab:tab3}.
(SSS: Sound Source Segmentation, AVC: Audio-Visual Caption, SSU-D: Sound-Source Understanding with Description Only, SSU-S: Sound-Source Understanding with Segmentation Only.)
The SSS yields similar segmentation indicators to our method; however, it only achieves basic segmentation and localization, lacking a deeper understanding of the scene. 
It is noticed that the AVIS mostly gains the second-best results in segmentation. However, AVIS overlooks exploring the interactive correlation between the given audio signals and text captions, leading to a shortage in temporal-wise caption production. Besides, we find that the AVSBench displays worse performance in metric mIOU. This is caused by the limitation in long-term spatial-temporal awareness extraction.
Due to differences in task and dataset formats, the AVC method produces unsatisfactory results for captioning, highlighting its limitations in fine-grained understanding. 
In contrast, SSUFormer demonstrates superior performance across all sub-tasks. The collaborative nature of the SSU tasks enhances both spatial and temporal perception, leading to improved overall understanding.

\subsubsection{Comparisons with multi-modal Large Models}
In recent years, multi-modal large models have developed rapidly~\cite{liu2024visual,han2023imagebind,wang2024modaverse,wu2023next}.
Some methods have also implemented architectures for multi-modality input and output, demonstrating significant versatility. As a task within multi-modality learning, SSU generates fine-grained spatial masks and frame-level captions. 
To evaluate the performance of multi-modality outputs using audio and video inputs, we select two state-of-the-art architectures: NExT-GPT~\cite{wu2023next}, ModaVerse~\cite{wang2024modaverse}, and PG-Video-LLaVA\cite{wang2024modaverse}.
Using LLMs\cite{zhang2023video} as the kernel of multi-modal large models enables the description of audio-visual data by simply modifying the output format without the need for additional training or fine-tuning.
However, despite advancements in visual representation learning, most methods still struggle with sound source localization in audio-visual contexts, even when guided by text prompts.
Our comparison focuses on how well these multi-modal large models understand audio-visual scenes, specifically in describing the sounding objects.
The results are reported in Tab.~\ref{tab:tab4}.
We use the same testing sets and evaluate the textual description results of the two datasets as a whole.
The NExT-GPT and ModaVerse are two MLLMs that could settle the audio-driven understanding.
The core idea of PG-Video-LLaVA is to effectively extend the capabilities of large-scale multimodal models based on images to the video domain, and endow them with pixel-level visual localization capabilities, while enhancing the understanding of video context by integrating audio signals.
Although this research can achieve similar functionality as our proposed method, similar to other large model based methods compared in the main paper, PG-Video-LLaVA does not perform well in our fine-grained data validation metrics. This also validates the difference between our method and the large model method.
Overall, our proposed SSUFormer exhibits superior task-specific abilities in audio-visual learning scenarios. This highlights the limitations of current multi-modal large models and emphasizes the need for further expansion of their generalization capabilities.
It is essential to note that the RA-SSU task's objectives differ from those of the large models; the comparison is made solely within the framework of multi-modality learning.

\begin{table}[t]
\centering
\caption{Comparison with multi-modal Large Models on sound object description task (average of two datasets).}
\label{tab:tab4}
\resizebox{0.48\textwidth}{!}{%
\setlength{\tabcolsep}{5.5 mm}
\begin{tabular}{lccc}
\toprule
Methods & BLEU & ROUGE-L & METEOR \\ \midrule \midrule
NExT-GPT~\cite{wu2023next}\textcolor[HTML]{C0C0C0}{$_{ICML'24}$} & 0.1207 & 0.1159 & 0.1569 \\
ModaVerse~\cite{wang2024modaverse}\textcolor[HTML]{C0C0C0}{$_{CVPR'24}$} & 0.1529 & 0.1476 & 0.1278 \\
PG-Video-LLaVA\cite{munasinghe2023pg}\textcolor[HTML]{C0C0C0}{$_{Arxiv}$} & 0.2300        & 0.2649           & 0.2260   \\ \midrule
RA-SSU Baseline & 0.2661 & 0.3102 & 0.2540 \\
\rowcolor[HTML]{ECF4FF} 
\textbf{SSUFormer} & \textbf{0.4676} & \textbf{0.5168} & \textbf{0.4479} \\ \bottomrule
\end{tabular}%
}
\end{table}

\begin{table}[h]
\centering
\caption{Quantitative comparisons on ablation analysis.}
\label{tab:tab5}
\resizebox{0.48\textwidth}{!}{%
\setlength{\tabcolsep}{3 mm}
\begin{tabular}{lccccc}
\toprule
\multicolumn{6}{c}{\textbf{Ablation Dataset: f-Lifescene}} \\ \midrule \midrule
Methods & mIoU & F-score & BLEU & ROUGE-L & METEOR \\ \midrule
Baseline & 0.4892 & 0.6510 & 0.1331 & 0.1285 & 0.1052 \\
+ Focal & 0.5150 & 0.6862 & 0.3201 & 0.2878 & 0.4531 \\
+ MCM  & 0.6132 & 0.6947 & 0.4429 & 0.4590 & 0.4274 \\
+ RCC  & 0.6141 & 0.7087 & 0.4621 & 0.4728 & 0.4452 \\\midrule
\rowcolor[HTML]{ECF4FF} 
\textbf{+ MoHE (SSU)} & \textbf{0.6530} & \textbf{0.7368} & \textbf{0.5290} & \textbf{0.5388} & \textbf{0.5050} \\ \bottomrule
\end{tabular}%
}
\end{table}

\subsubsection{Ablation Study}
To assess the effectiveness of the various modules and loss functions in the SSUFormer, we conducted a series of ablation analyses using the f-Lifescene dataset, which offers a diverse range of categories.
The quantitative results are shown in Tab.~\ref{tab:tab5} and the qualitative results are shown in Fig.~\ref{Figure11}.
Our findings indicate that each module design contributes positively to the final outcome metrics. With the stacking of modules and the introduction of additional loss functions, the SSUFormer ultimately outperforms all other configurations, achieving the best results.
Notably, the coordinated design of the Mask Collaboration Module and the Regional Consistency Constraint significantly enhances the model's understanding capabilities, demonstrating the importance of synergistic module interactions. Moreover, we observe that once we add the MoHE module, the performance of our framework displays a remarkable improvement (\eg, ROUGE-L: 0.4728 $\rightarrow$ 0.5388). This highly proves the effectiveness of our design on hierarchical prompting integrated with LLLaVA experts, which adaptively facilitates the framework to generate authoritative text descriptions in long sequence modeling.

\subsection{Qualitative Analysis}
\subsubsection{Experimental Results of SSUFormer}
\begin{figure}[h]
\centering
\includegraphics[width=0.49\textwidth]{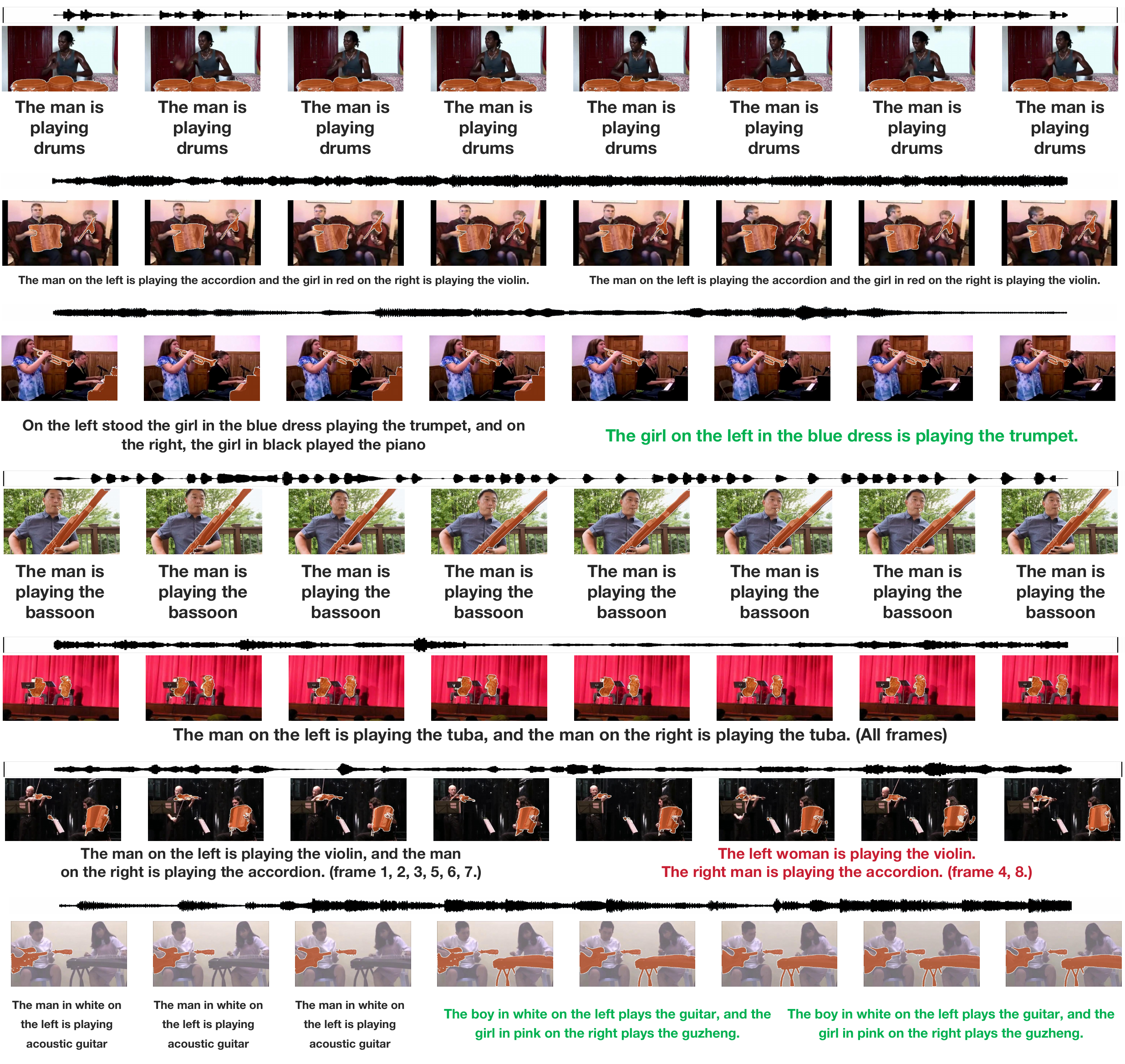} 
\caption{Experimental result samples of the SSUFormer on f-Music. The red-highlighted words mean unreasonable output. The green-highlighted words mean satisfactory output. (Zoom in for better details.)}
\label{Figure9}
\end{figure}

\begin{figure}[h]
\centering
\includegraphics[width=0.49\textwidth]{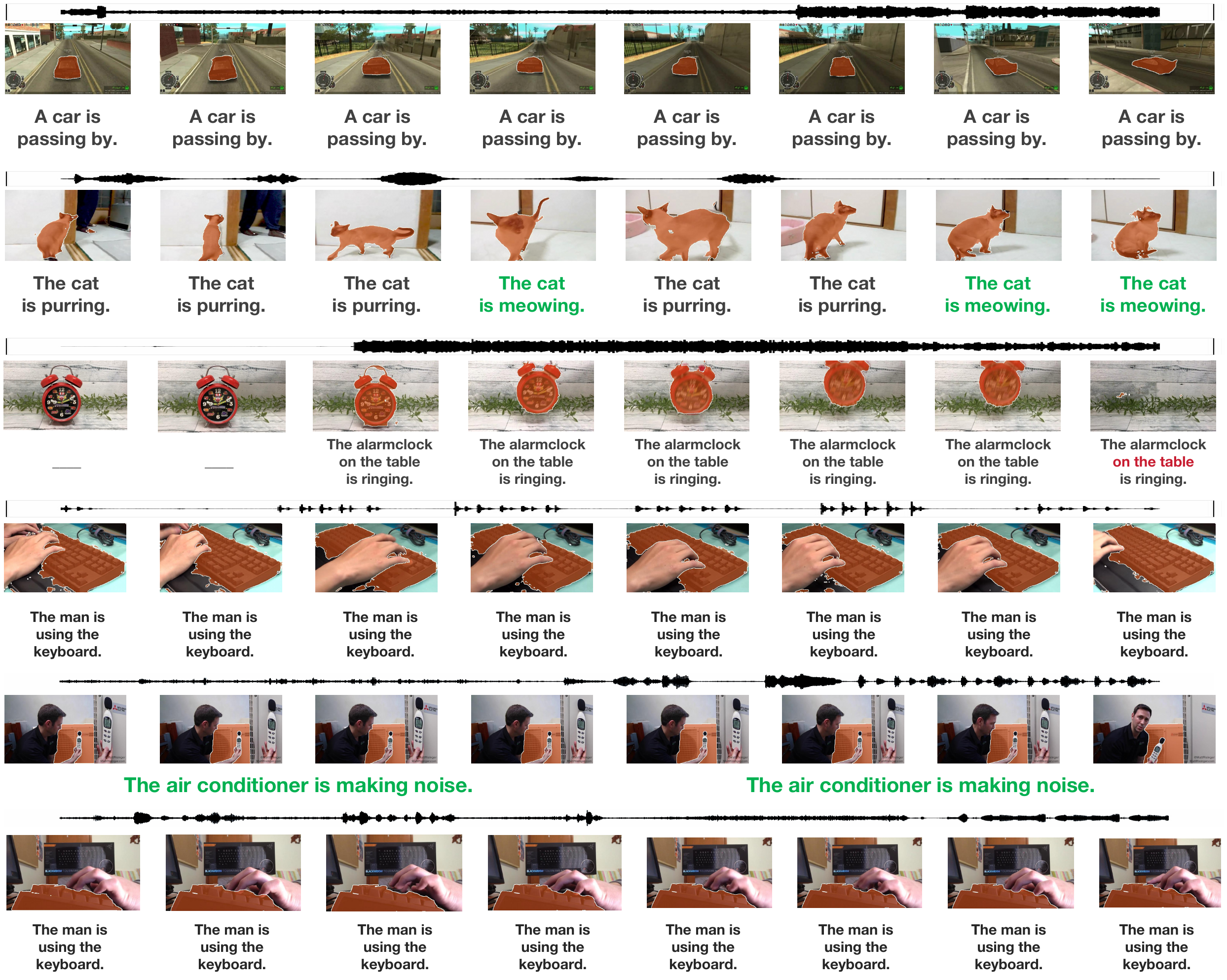} 
\caption{Experimental result samples of the SSUFormer on f-Lifescene. The red-highlighted words mean unreasonable output. The green words mean accurate perception. (Zoom in for better details.)}
\label{Figure10}
\end{figure}

To evaluate the performance of our method, we sampled four instances from two datasets randomly, and analyzed the result prediction, as shown in Fig.~\ref{Figure9} and Fig.~\ref{Figure10}.
For each sample, we extracted 8 images at one-second intervals from the middle segment of a ten-second clip.
Our method successfully achieves fine-grained sound object localization and description, aligning with our original dataset objectives. It effectively perceives multiple sound sources and understands dynamic sound scenes, as illustrated by the cat sample in Fig.~\ref{Figure10}.
Thus, we propose a region-aware, frame-level, and high-quality benchmark for sound source understanding, paving the way for future research in this area.
Meanwhile, we have also shown some complex results from the verification, such as multiple instruments sounding in different periods and complex life scenarios with uncommon perspectives, overlapping objects. (Line 2, 3, 6, 7 in Fig.9 and Line 5, 6 in Fig.10). 
These results demonstrate that our model is effective in complex scenes, especially in multi-object sound scenes.
However, some limitations persist. Inaccuracies remain in spatial localization and temporal description, as seen in the alarm clock data from the f-Lifescene dataset and the last sample in Fig.~\ref{Figure9} (red highlighted). 
Notably, the model continues to output the description ``on the table" even after the alarm clock leaves the field of view. Additionally, the temporal inconsistencies observed in Fig.~\ref{Figure9} need to be addressed.

\subsubsection{Ablation Illustration}
\begin{figure}[h]
\centering
\includegraphics[width=0.49\textwidth]{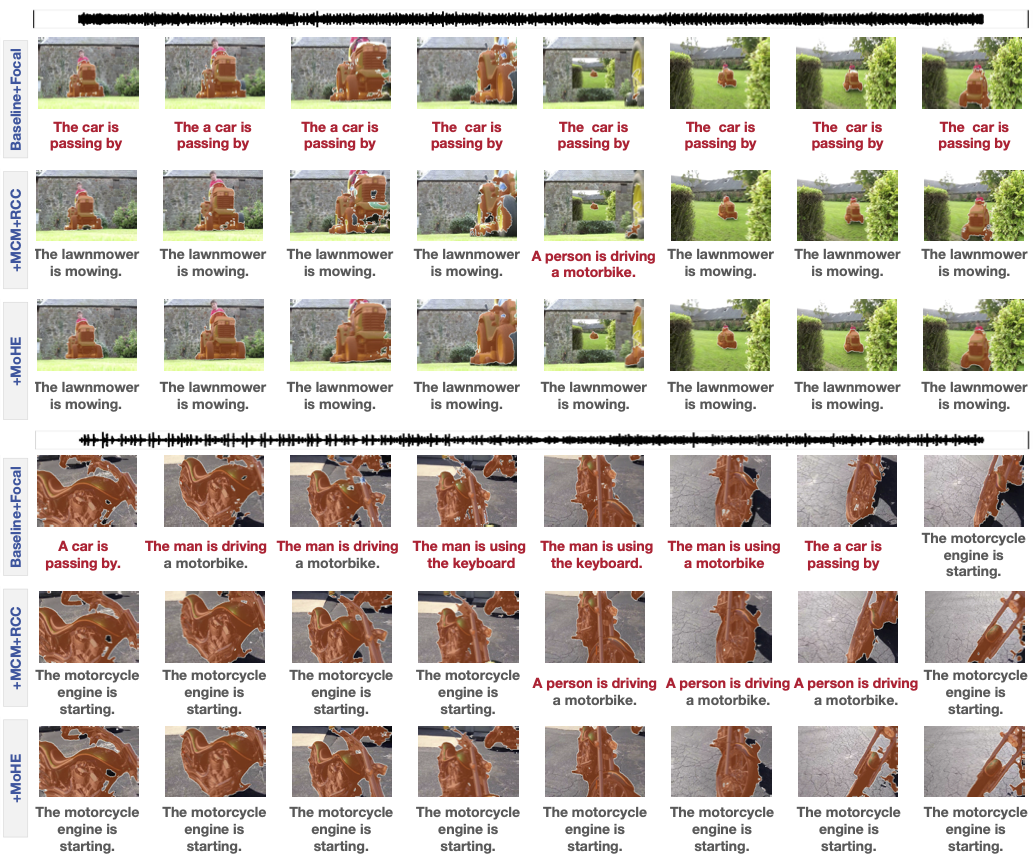} 
\caption{Data samples of the ablation analysis. Due to layout limitations, we present two groups of comparative results in f-Lifescene with more category diversity. MCM and RCC simultaneously improved these two task performances; MoHE further improves the caption results. The red-highlighted words mean unreasonable output. (Zoom in for better details.)}
\label{Figure11}
\end{figure}

Meanwhile, we have also sampled the results of the ablative experiments to verify the visualization effect of our proposed module. 
As illustrated in Fig.~\ref{Figure11}, the final SSUFormer (last row of each sample) provides more continuous, stable, and accurate results, which is also better in region detail.
Since the baseline model is relatively average performed, we mainly validate our two main new designs, MCM and RCC, as shown in Fig.~\ref{Figure11} (+MCM +RCC).
Both the segmentation and description performances have been improved. 
It is obvious that the Multi-Task Coordination Mechanism (MCM) enhances the clarity and consistency of descriptions used in multi-task settings.
Additionally, the Description Interaction in the Rational Contextual Coordination (RCC) further improves the effectiveness of segmentation. Together, these mechanisms significantly enhance the overall performance of the model in multi-task learning scenarios.
In addition, as depicted in the last row of Fig.~\ref{Figure11}, our framework incorporated with the MoHE produces coherent video captions \wrt audio signals and preserves temporal consistency in long sequences.
Meanwhile, we notice that our MoHE improves the segmentation results simultaneously. 
During model training, the better caption results encourage gradient backpropagation to focus on the generation of visual masks from another branch. This also validates our insight into the co-consistency between visual semantics and text in the design of MoHE. 

\subsubsection{Model Emergence for Other Tasks}
At last, as a multi-task, we discuss the benefits of the Sound (SSU) task for the audio-visual community, particularly its emergent capabilities. The most significant improvement in fine-grained SSU perception is its ability to assist in video retrieval and comprehension tasks.
Previous audio-visual retrieval tasks typically match data based on the overall features of audio and video. For instance, one might use an audio clip to retrieve a video of a ``dog barking" or employ a video clip to find the most similar video. However, this method allows only for overall feature matching.
In contrast, the SSU strategy enables us to search for or match specific regions within audio and video, such as a white dog barking continuously on the left side of the video. 
By matching these refined descriptions, we can achieve more accurate results and obtain frame-level predictions that pinpoint which clips in a video align with our retrieval objectives.
Moreover, our RA-SSU framework provides more detailed comprehension for video understanding tasks, offering emergent possibilities for future video learning-related tasks.

\section{Discussion}
In this paper, we propose a fine-grained audio-visual understanding framework. We aim to explore region-aware object localization and description in audio-visual scenes. 
This method is set to target task-specific settings, in which we can mainly solve audio-visual learning in specific scenarios, which is also the main difference from current Large Models such as large video models and general audio-visual models\cite{munasinghe2023pg, wu2023next, wang2024modaverse}. 
Therefore, our model is actually a specific transformer, using a specific dataset to achieve specific effects.

\begin{table}[h]
\centering
\caption{Computational Efficiency of the proposed Method with comparisons. (Params: model parameters. FPS: frame rate inferred per second. FLOPS: floating point operations per second. Inference Time: inference time for a single sample.)}
\resizebox{0.48\textwidth}{!}{
\begin{tabular}{lcccc}
\toprule
\multicolumn{1}{c}{\multirow{2}{*}{\textbf{Methods}}} & \multicolumn{4}{c}{\textbf{Computational Efficiency}} \\ \cmidrule{2-5} 
\multicolumn{1}{c}{} & \multicolumn{1}{c}{\textbf{Params}} & \multicolumn{1}{c}{\textbf{FPS}} & \multicolumn{1}{c}{\textbf{FLOPS}} & \multicolumn{1}{c}{\textbf{Inference Time}} \\ \midrule \midrule
\multicolumn{1}{c}{AVSBench\cite{zhou2022audio}}                         & \multicolumn{1}{c}{100.14M}      & \multicolumn{1}{c}{251. 45}    & \multicolumn{1}{c}{154.20G}           & \multicolumn{1}{c}{0.039}               \\ 
\multicolumn{1}{c}{AVIS\cite{guo2025audio}}                         & \multicolumn{1}{c}{527.3M}      & \multicolumn{1}{c}{23. 72}    & \multicolumn{1}{c}{2.65T}           & \multicolumn{1}{c}{0.435}               \\ 
\multicolumn{1}{c}{PG-Video-LLaVA\cite{munasinghe2023pg}}                         & \multicolumn{1}{c}{6.76B}      & \multicolumn{1}{c}{475.81}    & \multicolumn{1}{c}{1.32T}           & \multicolumn{1}{c}{0.635}               \\ 
\multicolumn{1}{c}{NExT-GPT\cite{wu2023next}}                         & \multicolumn{1}{c}{6.74B}      & \multicolumn{1}{c}{455.86}    & \multicolumn{1}{c}{1.62T}           & \multicolumn{1}{c}{0.834}               \\ 
\multicolumn{1}{c}{ModaVerse\cite{wang2024modaverse}}                   & \multicolumn{1}{c}{7.98B}      & \multicolumn{1}{c}{359.72}    & \multicolumn{1}{c}{1.70T}           & \multicolumn{1}{c}{0.658}                \\ \midrule
\rowcolor[HTML]{ECF4FF} 
\multicolumn{1}{c}{\textbf{RA-SSU (Ours)}}                         & \multicolumn{1}{c}{319.62M}      & \multicolumn{1}{c}{84.83}    & \multicolumn{1}{c}{1.596T}           & \multicolumn{1}{c}{0.117}               \\ \bottomrule
\end{tabular}}
\end{table}

For real-world applications, we have also compared the computational efficiency with different methods, as shown in Tab.VI.
The computational efficiency of our method is close to previous audio-visual learning methods.
It is more suitable for applications in specific small scenarios, used in environments that pursue perception accuracy and precision, such as the musical instrument scenes and daily life scenes.
Thus, our method cannot solve various types of open-set perception like large models. Due to the fewer parameters, this method is more convenient to deploy and apply.
The performance of the large models in specific scenarios is also not as good as our model, as shown in the experimental comparison Tab.IV.

There are also some limitations of our proposed method. As mentioned above, our method cannot settle the open-set perception.
This will limit the generalization of the model and limit our model from being fixed in special environments.
Meanwhile, this task-specific model is more sensitive to audio.
When the environment changes, especially when the audio information is new or brings noise, there will be deviations in the results. 
Furthermore, though our method could settle the fine-grained perception, the data acquisition process is quite complex. 
In real-world application scenarios, problems caused by data can increase the burden on project progress.

\section{Conclusion}
In this paper, we define a multi-modality learning task called Region-Aware Sound-Source Understanding (RA-SSU), which surpasses existing coarse-grained audio-visual learning (AVL) tasks. RA-SSU emphasizes the spatial localization and semantic understanding of sounding objects, enhancing the ability to scene understanding in complex scenarios.
To validate the RA-SSU task, we construct two datasets, f-Music and f-Lifescene. f-Music focuses on audio mixtures in music scenes, while the other covers complex everyday scenarios with interactions between multiple sounding objects. 
Moreover, we provide a benchmark method SSUFormer, a method that employs a multi-modal input and output Transformer architecture, featuring a specifically designed Mask Collaboration Module and a Regional Consistency Constraint. These components jointly improve the performance in sounding object segmentation and region description. 
Extensive experiments are conducted on our two datasets to verify the feasibility of the task, evaluate the availability of the datasets, and demonstrate the superiority of the SSUFormer.
Although multi-modal large models have developed rapidly, our method retains specific advantages in addressing challenges in particular scenarios.
In the future, we aim to expand the datasets with more categories, implement open-vocabulary learning scenarios, and enhance the universality of audio-visual tasks. Additionally, we will improve the continuity and consistency of results in the temporal dimension.


\bibliographystyle{IEEEtran}
\bibliography{Template}


 




\vfill

\end{document}